\documentclass{article}

\usepackage{arxiv}

\usepackage[utf8]{inputenc} 
\usepackage[T1]{fontenc}    
\usepackage{hyperref}       
\usepackage{url}            
\usepackage{booktabs}       
\usepackage{amsfonts}       
\usepackage{nicefrac}       
\usepackage{microtype}      
\usepackage{lipsum}		
\usepackage{graphicx}
\usepackage{natbib}
\usepackage{doi}
\usepackage{algorithm}
\usepackage{algorithmic}

\title{POGD: Gradient Descent with New Stochastic Rules}

\date{} 					

\author{ Feihu Han\\
    Deakin University\\
    \texttt{feihuh@deakin.edu.au} \\
    \And
    Sida Xing\\
    Deakin University\\
    \texttt{sxing@deakin.edu.au}
    \And
    Dr Sui Yang Khoo\\
    Deakin University\\
    \texttt{sui.khoo@deakin.edu.au}
}

\renewcommand{\headeright}{}
\renewcommand{\undertitle}{}
\renewcommand{\shorttitle}{\textit{POGD}}

\hypersetup{
pdftitle={POGD: Gradient Descent with New Stochastic Rules},
pdfauthor={Feihu Han, Sida Xing, Sui Yang Khoo},
}

\begin{document}
\maketitle

\begin{abstract}
	There introduce \emph{Particle Optimized Gradient Descent (POGD)}, an algorithm based on the gradient descent but integrates the particle swarm optimization (PSO) principle to achieve the iteration. From the experiments, this algorithm has adaptive learning ability. The experiments in this paper mainly focus on the training speed to reach the target value and the ability to prevent the local minimum. The experiments in this paper are achieved by the convolutional neural network (CNN) image classification on the MNIST and cifar-10 datasets.
\end{abstract}

\section{Introduction}
Gradient descant is a popular optimization in neural networks. Nowadays, neural network has attracted much attention in any fields, there are many different types of the neural networks that has already been developed\citep{haykin2009neural}. The convolutional neural networks is very popular in the classification field\citep{huang2017densely}, even though it is a kind of the feedforward neural network\citep{haykin2009neural}. In recent years, the structure of convolutional neural network has been rapidly improved\citep{smith2016deep}. However, many improvements develop the structural algorithm of convolutional neural network, but not much improve it from optimization. Generally, the optimization which is used for the common convolutional neural network is the gradient descent optimization. In fact, the method has the problem about local minimum and this problem is an issue for the model convergence\citep{bottou1991stochastic}.

Here proposes \emph{Particle Optimized Gradient Descent}, a method for gradient descent with time memory. This algorithm can through iteration history to find the expect global minimum and achieve the convergence. It has fast convergence speed, adaptive learning ability, good global optimizing ability. POGD also can improve learning outcomes and prevent the local minimum in some extent for the convolutional neural networks’ gradient descent processing.

\section{Application}
The application presents in here is image classification. In the current neural networks’ frames, convolutional neural network has good processing effect for the image classification and processing\citep{lecun2010convolutional}. The mainly processing of the convolutional neural network is in hidden layers\citep{lecun1998gradient}. Although the structure of convolutional neural network is constantly evolving, convolution and pooling are still its core algorithms since the first generation\citep{lecun1998gradient}. The image identification for neural networks is targeted, there need to use the certain database to train the neurons to classify the certain feature from unknown pictures\citep{lecun2010convolutional}. The traditional neural network needs to through a large dataset and long-term training to achieve the image identification for a certain training\citep{haykin2009neural}. In contrast, although convolutional neural network still requires a large number of training sets, it can extract image features and perform image classification better\citep{lecun1998gradient}. Under the same conditions, the convolutional neural network have better results than the traditional neural network\citep{mcculloch1943logical}. The experiments here uses Alexnet\citep{krizhevsky2017imagenet} as the basic model to compare the differences between algorithms. The hidden layer for this AlexNet includes convolutional layer, maximum pooling layer and activation layer\citep{krizhevsky2017imagenet}. ReLU\citep{kuo2016using} is used as the activation function.

\section{Relative Research}
To achieve the \emph{Particle Optimized Gradient Descent}, there need to introduce the relevant optimization algorithms include stochastic gradient descent\citep{darken1991towards}, AdaGrad\citep{lydia2019adagrad}, Adam\citep{duchi2011adaptive} and particle swarm optimization\citep{kennedy1995particle}. This algorithm still belongs to the gradient descent algorithm, therefore, the development of traditional gradient descent algorithms is introduced.

\subsection{Batch Gradient Descent}
\begin{equation}
	\theta_{t} = \theta_{t-1} - \eta\nabla_{\theta_{t-1}}E(\theta_{t-1})
\end{equation}
Batch gradient descent is a basic method to introduce the gradient descent in the neural network. As it shows in Equation 1\citep{ruder2016overview}, for batch data “$\theta$”, there calculate the gradient “$\nabla _\theta$” in form E($\theta _{t-1}$). As a standard method, there only subtract the value from the corresponding derivative to complete the iteration. For the large datasets, batch gradient descent has redundant computations when it recalculates the gradients for similar samples\citep{ruder2016overview}.

\subsection{Stochastic Gradient Descent}
\begin{equation}
    \theta_{t} = \theta_{t-1} - \eta\nabla_{\theta_{t-1}}E(\theta_{t-1},x_{i})
\end{equation}
Stochastic gradient descent updates the parameter for each input rather than whole training batch\citep{ruder2016overview}. In the stochastic gradient descent shows in Equation 2\citep{darken1991towards}, "$\eta$" is the learning rate, "x$_i$" is a random sample is chosen from the batch\citep{darken1991towards}. Stochastic gradient descent eliminates redundant computations in batch gradient descent by performing only one gradient update at a time. When stochastic gradient descent performs iterations, the objective function will fluctuate within a larger variance range than batch gradient equation. Through this feature, if the iteration gets into the local minimum, the fluctuation of the objective function is able to jump out of the current local minimum and find a better position. On the other hand, the greater variance fluctuation will cause convergence process more difficult. Adjusting the learning rate and decreasing the learning rate as the iteration increases can help the algorithm achieve better convergence effect\citep{ruder2016overview}.

\subsection{AdaGrad}
\begin{equation}
    g_{t,i} = \nabla_{\theta_{t}}E(\theta_{t,i})
\end{equation}
\begin{equation}
    G_{t,i} = G_{t-1,i} + g_{t,i} * g_{t,i}
\end{equation}
\begin{equation}
    \theta_{t,1} = \theta_{t-1,i} - \frac{\eta}{\sqrt{G_{t,ii}+\epsilon}} * g_{t,i}
\end{equation}
\begin{equation}
    \theta_{t} = \theta_{t-1} - \frac{\eta}{\sqrt{G_{t}+\epsilon}} * g_{t}
\end{equation}
AdaGrad adapts learning rate to update the aim value. Previous algorithm uses one learning rate for all parameters in the same iteration. However, as it shows in Equation 3\citep{lydia2019adagrad}, AdaGram uses different learning rate for each of the parameters in the same iteration. In AdaGrad, it does larger iteration for infrequent parameter and does smaller iteration for frequent parameter\citep{ruder2016overview}.

Equation 4\citep{lydia2019adagrad} shows the calculation of the most important parameter “G${_t,i}$”. “G${_t,i}$” is a diagonal matrix for each diagonal element “i”. This parameter iterates by using the Hadamard product square of the gradient of corresponding aim value. Equation 5\citep{lydia2019adagrad} shows the iteration of the aim value $\theta_{_t,i}$. There calculates the learning rate with the quotient of the arithmetic square root of “G$_t$”, then calculates the Hadamard product of the corresponding value “g$_{t,i}$” and the final product is used to complete the iteration. Constant “$\epsilon$” is an value to prevent void value from “G$_{t,i}$”\citep{lydia2019adagrad}. Generally, it set as a very small number 10$^{-8}$\citep{ruder2016overview}.

Owing to “G$_t$” includes the sum of the squares of the past gradients for all parameters in the same iteration, as it shows in Equation 6\citep{ruder2016overview}, there can use an element-wise matrix-vector multiplication between “G$_t$” and “g$_t$”\citep{ruder2016overview}.

\subsection{Momentum}
\begin{equation}
    \triangle\theta_{t} = -\eta\nabla_{\theta_{t}}E(\theta_{t}) + p\triangle\theta_{t-1}
\end{equation}

\begin{equation}
    v_{t} = p\triangle\theta_{t-1} + \eta\nabla_{\theta_{t}}E(\theta_{t})
\end{equation}

\begin{equation}
    \theta_{t} = \theta_{t-1} - v_{t}
\end{equation}

Momentum gradient descent introduce the momentum from physics into the gradient descent. As it shows in Equation 7\citep{qian1999momentum}, on the basis of general gradient descent, there gives the product of a momentum parameter “p” and the aim value at previous time as a new bias to update the aim value. Current aim value $\theta_t$ is updated by the momentum from last aim value $\theta_{t-1}$ and gradient value in corresponding time. The momentum parameter “p” generally is a constant 0.9\citep{qian1999momentum}. Owing to the coefficient is less than 1, this allows the entire integration to automatically achieve convergence as the number of iterations increases\citep{ruder2016overview}. Stochastic gradient descent is difficult to jump out of the local minimum under more complex conditions. The introduction of momentum also makes gradient descent has further improved for the local minimum problem\citep{ruder2016overview}.

In the neural network iteration, generally uses the Equation 8\citep{ruder2016overview} and Equation 9\citep{ruder2016overview} to process the momentum gradient descent\citep{ruder2016overview}. These two equations combine the gradient values and according to the basic gradient descent method, there use the difference between the target value and the gradient to calculate the needed new target value.

\subsection{Adam}
\begin{equation}
    g_{t} = \nabla_{\theta_{t}}E(\theta_{t,i})
\end{equation}

\begin{equation}
    m_{t} = \beta_{1}*m_{t-1} + (1-\beta_{1})*g_{t}
\end{equation}

\begin{equation}
    v_{t} = \beta_{2}*v_{t-1} + (1-\beta_{2})*g^{2}_{t}
\end{equation}

\begin{equation}
    \theta_{t} = \theta_{t-1} - \frac{\eta*m_{t}}{\sqrt{v_{t}}+\epsilon}
\end{equation}

Adam is a popular adaptive gradient descent algorithm. In the above equations, “$\theta$” is data and “g” is the gradient of the data. Equation 10\citep{duchi2011adaptive} shows the calculate gradient for each input data. Equation 11\citep{kingma2014adam}, Equation 12\citep{kingma2014adam} and Equation 13\citep{kingma2014adam} are the main update of Adam. For these equations, “m” stands for the first momentum estimate and “v” is the second momentum estimate. Value "$\eta$" is the step size or learning rate. Constant “$\beta_1$” is the exponential decay rate for the first momentum estimate. Constant “$\beta_2$” is the exponential decay rate for the second momentum estimate. There is a constant “$\epsilon$” is used to add in noise to enhance the generalization performance and prevention overfitting\citep{kingma2014adam}. Generally, “$\beta_1$” is set as 0.9, “$\beta_2$” is set as 0.999, “$\epsilon$” is set as 10$^{-8}$\citep{kingma2014adam}.

From the Adam equations, Adam can calculate the adaptive learning rate of each params. It integrates the exponential decaying average of squared gradient and exponential decay average. As a new algorithm that integrate the advantages of other algorithms, it has faster convergence and better training than other adaptive learning rate algorithms\citep{kingma2014adam}.

\subsection{Particle Swarm Optimization}
Particle swarm optimization is an optimization algorithm developed from the research of natural swarm such as bird. Based on the foraging patterns of bird swarm, the algorithm has randomness and ability to simulate social swarm. particle swarm optimization is related to evolutionary algorithm and genetic algorithm. Through the iteration of individual particles and particle groups, this algorithm is able to obtain the expected value from the continuous nonlinear functions. The further research shows this algorithm can optimize the continuous nonlinear functions\citep{kennedy1995particle}.

\begin{equation}
    v_{n+1} = v_{n}+c_{1}r_{1}(Pbest-x_{n})+c_{2}r_{2}(Gbest-x_{n})
\end{equation}

\begin{equation}
    x_{n+1} = x_{n}+v_{n+1}
\end{equation}

\begin{equation}
    v_{n+1} = wv_{n}+c_{1}r_{1}(Pbest-x_{n})+c_{2}r_{2}(Gbest-x_{n})
\end{equation}

Equation 14 and Equation 15 is the functions to achieve TLC model. Equation 16 proposes the inertia factor on the original basis. For these functions, “x” is position and “v” is velocity. The inertia factor is presented by “w” with non-negative domain. The larger of “w”, the global seeking ability becomes stronger, but the local seeking ability becomes weaker. The smaller of “w”, the global seeking ability becomes weaker, but the local seeking ability becomes stronger\citep{shi1998modified}. Stochastic factors\citep{kennedy1995particle} are presented by “c”. The local best position is presented by “Pbest” and swarm best position is presented by “Gbest”. Therefore, it is already shown in Equation 11, the trust parameter “c$_1$” stand for the confidence for the current local particle position, another trust parameter “c$_2$” stand for confidence for the swarm position\citep{venter2003particle}. From the research, there can get good results when c$_1$ = c$_2$ = 2\citep{kennedy1995particle}. The “r” is a random number with domain [0, 1]\citep{shi1998modified}.

\subsection{Algorithm Directions}
Traditional gradient descent optimization usually starts from the error function of a given neural network, and this error function is parameterized by the weights in the neural network. The gradient descent method computes the gradient of the error function relative to each function, then it modifies the weight along the descent direction of the gradient to reduce the error\citep{qian1999momentum}. This method has problem about it can have a result from local minimum value but not suit for the overall situation\citep{bottou1991stochastic}.

From the concept of the particle swarm optimization, the algorithm is included into local optimization and global optimization.\citep{kennedy1995particle} Therefore, there is going to add this concept into the neural network to solve the local minimum problem\citep{bottou1991stochastic} from traditional gradient optimization algorithm.

\emph{POGD} uses the particle swarm optimization to optimize the weights of the convolutional neural. Weights have different values and significance in the neural network\citep{haykin2009neural}. There include the expected weights in the initialization phase, the neuron weights that used for learning in each layer, the feature value of the classifier in fully connection layers\citep{haykin2009neural}.

Through the backpropagation principle, to reduce the error between expected value and learned value can speed up the convergence\citep{lippmann1994book}. The research of iteration is to better update the expect value. Therefore, if there can successfully get more effective initialized weights, it should be possible to speed up the acquisition efficiency of feature weights. In this condition, the structure of the neural network will be simpler\citep{smith2016deep}, thereby achieving structural progress.

From the existing research, the particle swarm optimization is preferred to add in the backpropagation\citep{sinha2018particle,yamasaki2017efficient}.The particle swarm optimization can be used for the discontinuous nonlinear problems in neural networks\citep{kennedy1995particle}. For POGD, it introduce the particle swarm optimization concept into the traditional gradient descent optimization, it is successfully combined the non-linear advantage from PSO and the linear advantage from gradient descent together.

\section{Algorithm Description}
\subsection{Algorithm Equations}
\begin{equation}
    g_{t}=\nabla_{\theta_{t}}E(\theta_{t})
\end{equation}

\begin{equation}
    a_{t}=a_{t-1}+g_{t}*g_{t}
\end{equation}

\begin{equation}
    gb_{t} = \frac{g_{t}}{\sqrt{a_{t}}+\epsilon}
\end{equation}

\begin{equation}
    v_{t} = \omega m_{t-1}+c_{1}r_{1}(gb_{t-1}-g_{t})+c_{2}r_{2}(pb_{t-1}-g_{t})
\end{equation}

\begin{equation}
    m_{t}=-v_{t}
\end{equation}

\begin{equation}
    pb_{t} = g_{t}
\end{equation}

\begin{equation}
    \theta_{t}=\theta_{t-1}-\eta(g_{t}+v_{t})
\end{equation}

\emph{Particle Optimized Gradient Descent} is the proposed gradient descent algorithm for stochastic optimization. Equation 17 to 23 are the equations to calculate each parameter in POGD. Equation 17 is a general gradient calculation. Equation 18 and 19 is similar to AdaGrad. There calculates the Hadamard product “a$_t$” of the derivative of the corresponding aim value first. Then there calculates the quotient between gradient and arithmetic square root of “a$_t$”, this result is set to the global best value “gb$_t$”.

Equation 20 shows the main moment vector update in POGD. This function has same formation about progressive PSO velocity equation. This equation has different interpretation in this algorithm. Equation 21 is used to turn “v$_t$” to opposite sign. Equation 22 shows particle best “pb$_t$” is equal to gradient in last time point. As Equation 20 shows the second moment vectors is updated by the “m$_{t-1}$”, “gb$_{t-1}$” and “pb$_{t-1}$” Finally, Equation 23 shows the aim value is iterated by reduce the sum of current gradient and “v$_t$”.

The coefficients for the above equation in current setting are “$\epsilon$” = 10$^{-8}$, “$\omega$” = 0.9, “c$_1$” = 2, “c$_2$” = 1, “$\eta$" is the learning rate, “r$_1$” and “r$_2$” are random numbers with domain [0, 1]. According to the experiments, POGD algorithm supports the adaptive learning rate adjustment of the algorithm itself.

\subsection{Algorithm Updating}

\begin{algorithm}[H]
    \caption{Particle Optimized Gradient Descent}
    \begin{algorithmic}
        \REQUIRE $\eta$: Step size
        \REQUIRE $f(\theta)$: Stochastic objective function with parameters $\theta$
        \REQUIRE $\theta_0$: Initial parameter vector
        \STATE $a_0$ $\leftarrow$ 0 (Initialize $1^{st}$ moment vector)
        \STATE $v_0$ $\leftarrow$ 0 (Initialize $2^{nd}$ moment vector)
        \STATE $m_0$ $\leftarrow$ 0 (Initialize $3^{rd}$ moment vector)
        \STATE $gb_0$ $\leftarrow$ 0 (Initialize $4^{th}$ moment vector)
        \STATE $pb_0$ $\leftarrow$ 0 (Initialize $5^{th}$ moment vector)
        \WHILE{$\theta_{t}$ not converged}
            \STATE $g_{t}$ $\leftarrow$ $\nabla_{\theta_{t}}$E($\theta_{t}$) (Get gradient w.r.t. stochastic objective at timestep t)
            \STATE$a_{t}$ $\leftarrow$ $a_{t-1}$+$g_{t}$*$g_{t}$ (Update biased first moment estimate)
            \STATE$gb_{t}$ $\leftarrow$ $\frac{g_{t}}{\sqrt{a_{t}}+\epsilon}$ (Update global best value)
            \STATE$v_{t}$ $\leftarrow$ $\omega m_{t-1}+c_{1}r_{1}(gb_{t-1}-g_{t})+c_{2}r_{2}(pb_{t-1}-g_{t})$ (Update biased second moment vector estimate)
            \STATE$m_{t}$ $\leftarrow$ -$v_{t}$ (Update the second moment vector to opposite sign)
            \STATE$pb_{t}$ $\leftarrow$ $g_{t}$ (Update particle best value)
            \STATE$\theta_{t}$ $\leftarrow$ $\theta_{t-1}$ - $\frac{\eta*m_{t}}{\sqrt{v_{t}}+\epsilon}$
        \ENDWHILE
        \RETURN{$\theta_{t}$ (Resulting parameters)}
    \end{algorithmic}
\end{algorithm}

Above workflow shows the processing of the POGD. During the whole processing, it needs to use the parameters under correct time “t” according to the formular and the setting of each coefficient.

The original velocity equation in the particle swarm optimization ingeniously uses the local and global optimum to update particle velocities in non-linear function. This is feature that need to add into the gradient descent. Although gradient descent is linear problem, adding in the stochastic factors can make the ability for convergence jump out of the local minimum become stronger. As same as the particle swarm optimization, particle best position and global best position are the important parameters in this algorithm. However, this project integrates some concept from particle swarm optimization into the gradient descent, the traditional method of acquiring best location from iteration of expected function is abandoned. Therefore, fitness value\citep{chow2004surface} is updated by follow the equations rather than the traditional value comparison.

The important value “v” of this algorithm is used to increase the bias value to the gradient under “t” iteration to achieve the gradient descent. From the particle swarm optimization equations, the three parts to sum up as “v” are the historical “v” value; the calculation of the recorded global best “x” position and the current “x” value; the calculation of the recorded local best “x” position and the current “x” value. In this project’s integrate concept algorithm, these three parts are still retained. However, the update of the two best position value needs to be completed without using objective function in gradient descent. Through research, the current global best position “gb” is obtained by calculating the gradient coefficient of the iterative equation from AdaGrad method. The current local best position “pb” is equal to the gradient on last time point. There need to notice the historical value of “v” in this algorithm need to take the negative value of the previous time point. Therefore, for certain time “t”, historical value “m$_t$” has the opposite sign with “v$_t$”. In practice, there find it is easy for “v” to iterate uncontrollably in the direction of a certain extreme value when “v” is completely dependent on the positive superposition. To solve uncontrolled problem, -1 is used to multiply with “v$_t$” and get an opposite sign value “m$_t$”. Through this way, “v” can be updated in both positive and negative direction, rather than in one direction for the entire iteration. In the process of research, it has found that if “v” is allowed to be updated only in a single direction, it is easy to cause the neural network cannot converge or loss value grows uncontrollably to “nan”. “$\omega$” is the coefficient of “m”, it is used to control the effect of “m” from the last time on the current update. Generally, it is necessary to ensure that “m” can have sufficient influence on the current update. At the same time, with the overall iteration, the relevant parameters which are affecting aim value need to be reduced to update themselves to help the model achieves convergence. Therefore, in this project, “$\omega$” is set as constant 0.9.

In this algorithm, the objective function and swarm particle calculation in particle swarm optimization are replaced by directly predict through existing values. The expect value from last time point minus current time point gives a bias about the best solution to “v”. The previous time gradient minus current time point gives the bias to prevent excessive update. Through the research study of the algorithm, the global best bias is a method for the main iteration of “v” value. The calculation of this bias causes the algorithm to demonstrate an ability to accelerate the weight convergence, but the non-controlling results of extreme value and single direction value can cause the convergence iterates in the opposite side fast. Therefore, in addition to the negative direction value “m” is used in “v” updating, local best is also used to adjust the global best parameter. The calculation of local best value assumes that the previous position is the best value. Both “m” value and “pb” value is able to prevent excessive updating, but the calculation of “pd” makes updating of “v” has better local optimizing ability. Also, adjusting on “pb” is easier than on “m.”

According to the Equation 16, without considering the new method of assignment for each key parameter, the main adjustments will focus on each coefficient. In this project, “$\omega$”, “c$_1$” and “c$_2$” are the constants that obtained by research on the corresponding values of the original particle swarm optimization equation. The coefficient “$\omega$” of last state value has introduced in the above content. This coefficient generally assigns between 0 to 1. A larger value can emphasize the effect of the previous moment on the current update. Comparing with the global best calculation method, the subtraction of local best is larger than the global best. Therefore, to make the important global optimization has more important effect in the iteration, smaller coefficient is set for local best calculation. In the new algorithm proposed in this project, global best coefficient “c$_1$” is equals to 2, local best coefficient “c$_2$” is equals to 1. Small local best coefficient will lead to a decrease in the overall resistance to extreme values, large local best coefficient will cause iteration becomes prone to get stuck in local minimum. For the global best coefficient, larger value will emphasize global effect, but it needs to pay attention to if convergence is normal.

\subsection{Algorithm Features}

\begin{figure}[!h]
	\centering
	\includegraphics[width=6in]{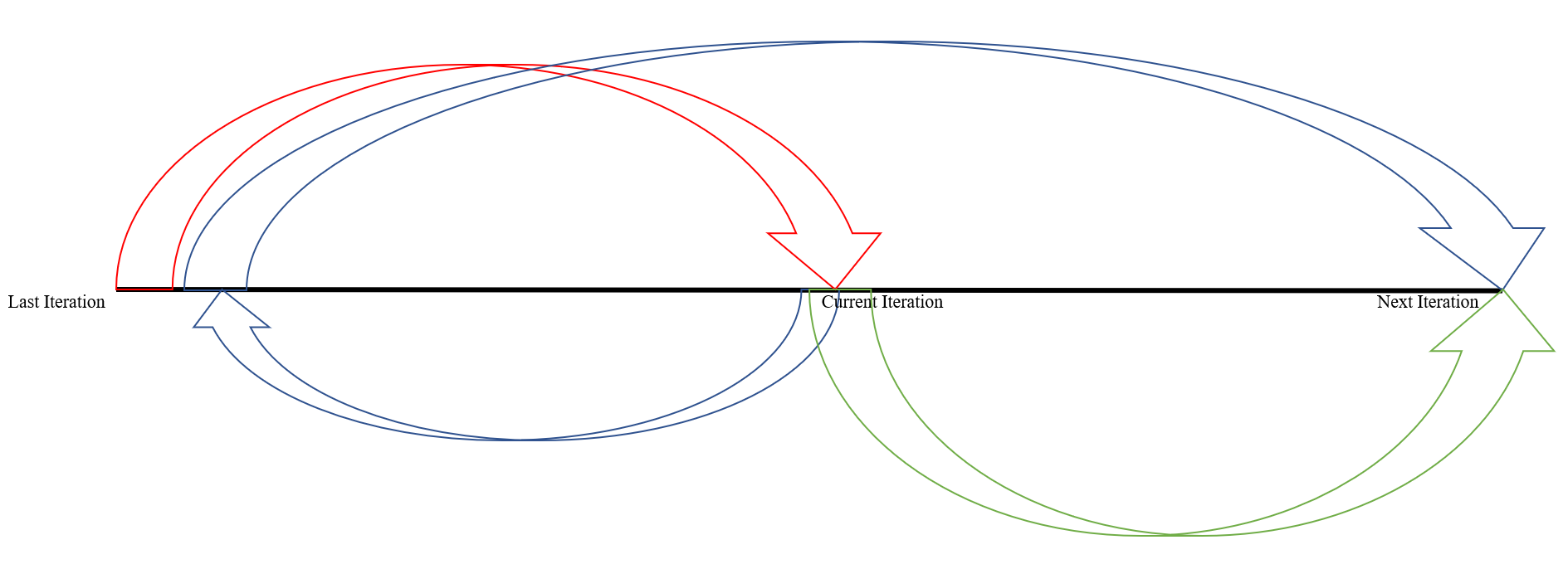}
	\caption{POGD Iteration Concept}
	\label{fig_IC}
\end{figure}

When calculating the current state, POGD allows controllable return to the vicinity of the origin for iterative calculation. The key point is the reason about current “v” need minus the value from last iteration. Figure \ref{fig_IC} shows the iteration concept of POGD. Blue arrows stand for the POGD iteration process. From the role of the process of velocity variable “v” in PSO position iteration formula, the velocity “v” is closer to the physical displacement concept in its mathematics expression. Assumption from this point, POGD makes the previous location becomes a part of basis to predict the next location. Owing to the gradient decent is essentially a linear problem, it has been found the accumulating displacement can easily make iteration loss convergence. POGD jump away from the current position and returns to the previous state, then combine the current iteration data to realize a good prediction of next position. By increase the value of “$\omega$”, the previous position has a greater effect on the current speed. When “$\omega$” is too small, the iteration of velocity value is easy to be affected by large step and cannot converge. This processing makes the result use initial value as the core but can be optimized by the historical gradient iteration.

The POGD velocity iteration seems like a kind of assumption about PSO when there is only one particle during the calculation. AdaGrad gradient iteration is used as a purposed function for global best value. Owing to there is only one particle in each iteration, so there uses the last gradient as the local best value. This is the main feature of this algorithm which is by using the optimal ability of PSO to improve gradient descent.

In traditional gradient descent, each iteration calculates from the previous result, it with poor global optimization ability and easy to get into a local minimum. POGD is affected by earlier information and integrate the PSO velocity iteration, it is also affected by the prediction of the gradient from current time and the gradient from previous time. Therefore, it has better global optimization ability.

\section{Experiments}
There set up two datasets with CNN image classification experiments. These two experiments show the features of POGD about convergence speed, adaptive learning ability and ability to prevent the local minimum.

\subsection{First Experiment: Mnist}
The first experiment uses the MNIST datasets and trained by a convolutional neural network model on NumPy. It is a handwritten digit recognition data base and has 60000 samples. It includes the handwritten images for Arabic numerals 0 to 9. These data are black-and-white images, so it only has single colour channel\citep{nielsen2015neural}. Therefore, for the primary model condition, the stride of the convolution processing and the maximum pooling processing in channel length is 1. Each of the image with 28 × 28 pixels size\citep{radford2015unsupervised}. The classifier is set on the 864 dimension image vectors. There uses mini-batch size of 64 and each of epoch has 936 steps. There are 20 epochs for each of the method and compares the results. 50\% dropout noise is applied to the neural network.

\begin{figure}[!h]
	\centering
	\includegraphics[width=3in]{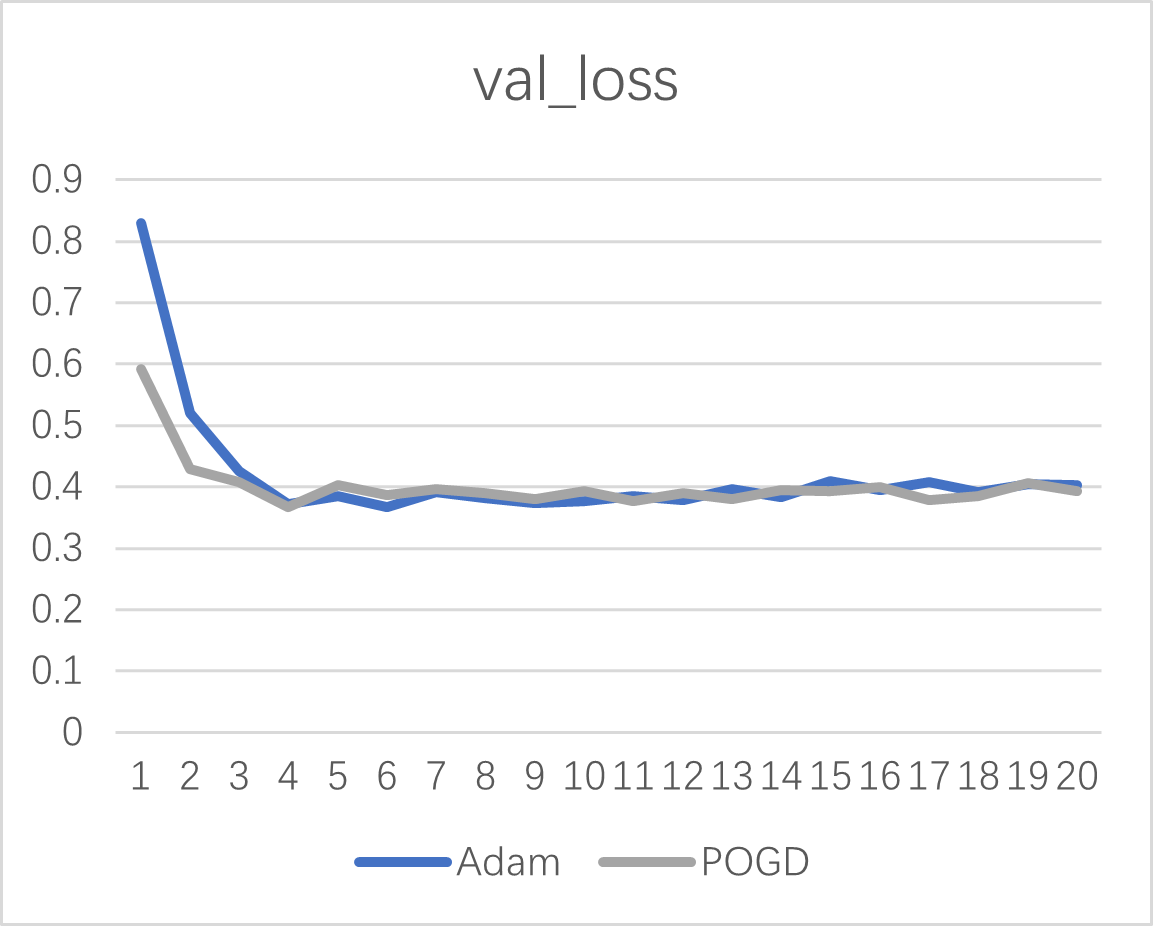}
	\caption{Validation Loss Curve for Mnist Dataset}
	\label{fig_VLM}
\end{figure}

\begin{figure}[!h]
	\centering
	\includegraphics[width=3in]{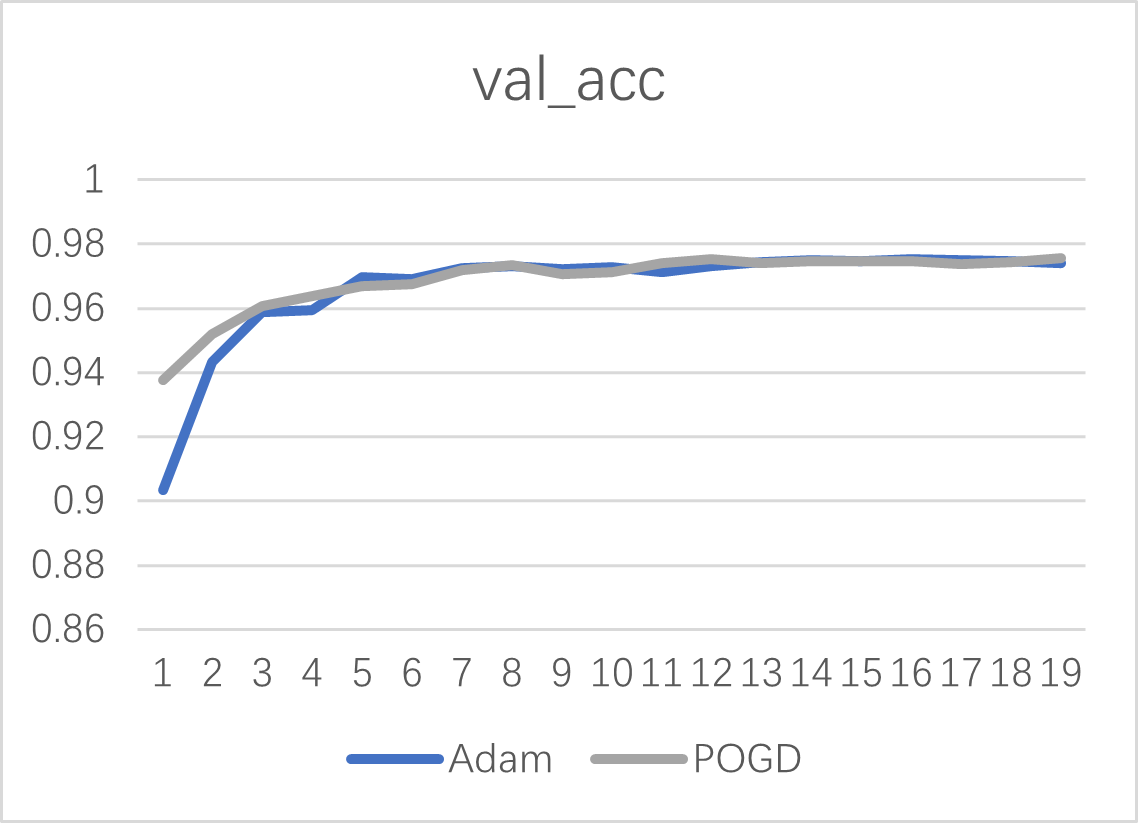}
	\caption{Validation Accuracy Curve for Mnist Dataset}
	\label{fig_VAM}
\end{figure}

Figure \ref{fig_VLM} shows both algorithms in the Mnist dataset’s validation loss together without the first epoch. Figure \ref{fig_VAM} shows both algorithms in the Mnist dataset’s validation accuracy together without the first epoch. Through the comparison, POGD has obvious smaller validation loss, and it is the fastest loss convergence speed for Mnist dataset. From Appendices, Figure \ref{fig_PLIM}, \ref{fig_ALIM} shows each iteration training loss changing for three models in first experiment. From the iteration changing curves, convergence of POGD algorithm is around 300 iterations faster than Adam. Therefore, in the beginning few epochs, POGD has better result. In the first experiment, the learning rate is reduced by the increases of epoch to help each algorithm can have better converge. POGD has better performance than Adam, it uses less iteration to achieve training and has better result.

From the validation loss results for Adam, the first epoch result is 0.83, it achieves 0.42 on the third epoch. From the validation loss results for POGD, the first epoch result is 0.59, it achieves 0.40 on the third epoch. To calculate the average value in the steady state, both of them is around 0.39. The validation accuracy curve can also find POGD rise faster. Therefore, in the same conditions, POGD has faster learning speed than Adam.

\subsection{Second Experiment: Cifar-10}
The second experiment uses the Cifar10 datasets and trained by a convolutional neural network model on Keras\citep{ketkar2017introduction}. It is a data base includes 60000 tiny colour images. It sets up 6000 examples in each of 10 classes. These data are colourful images, so it has three colour channels\citep{wang2016cnn}. For the primary model condition, the stride of the convolution processing and the maximum pooling processing in channel length is 3. Each of the image with 32 × 32 pixels size\citep{krizhevsky2010convolutional}. The last fully connection layer has 5130 parameters then classifies into 10 categories. There uses minibatch size of 32 and each of epoch has 50000 steps. There are 50 epochs for each of the method and compares the results. 50\% dropout noise is applied to the neural network. The learning rate is set as a constant 0.01 in the training model, but as the adaptive algorithm, POGD uses the same adaptive learning rate descent formula as Adam. This experiment is also used to test the adaptive learning rate of POGD. Batch normalization is used in the hidden layer to process the result from affine transformation and use the standardized results to train neural network parameters\citep{ioffe2015batch}. It uses the normalization process, the neuron weight of each hidden layer falls into the sensitive region of the activation function and avoids the vanishing gradient\citep{kuo2016using}.

\begin{figure}[!h]
	\centering
	\includegraphics[width=3in]{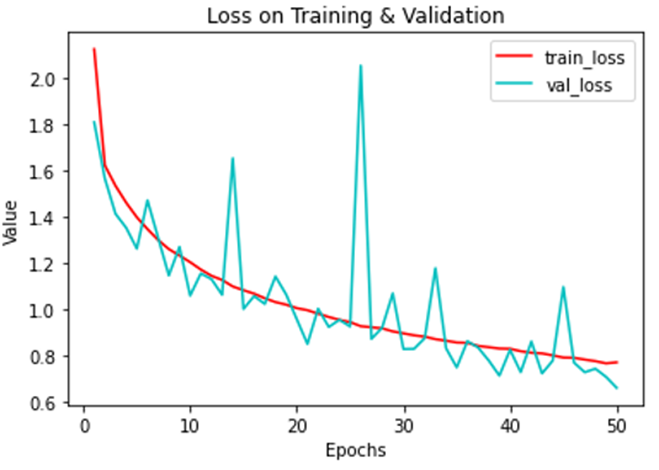}
	\includegraphics[width=3in]{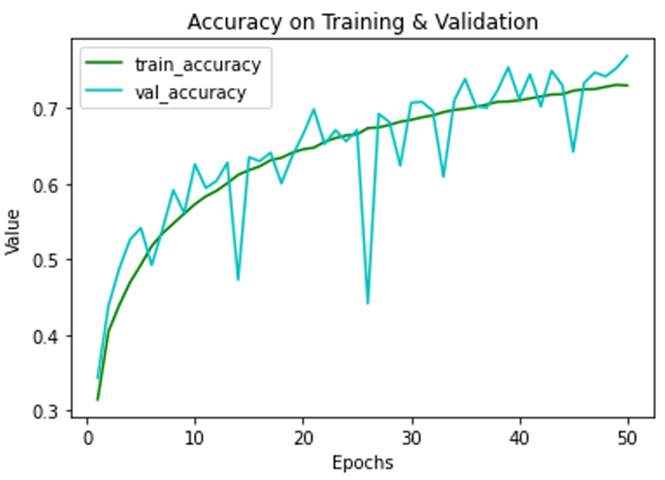}
	\caption{POGD Curves for Cifar10 Dataset}
	\label{fig_PCD}
\end{figure}

\begin{figure}[!h]
	\centering
	\includegraphics[width=3in]{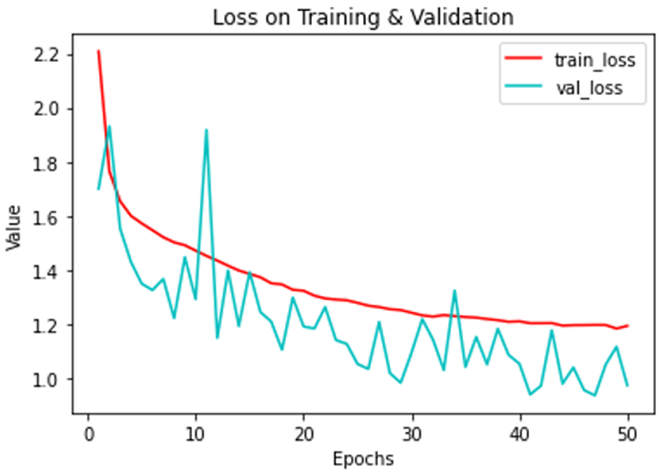}
	\includegraphics[width=3in]{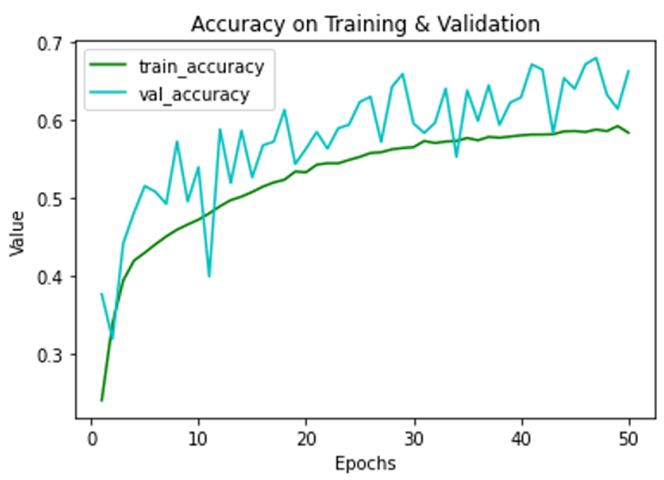}
	\caption{Adam Curves for Cifar10 Dataset}
	\label{fig_ACD}
\end{figure}

From the training set results of POGD model, the loss is 2.1262 for the first epoch and 0.7688 for the fiftieth epoch. The accuracy is 0.3148 on the first epoch and 0.7302 on the fiftieth epoch. From the validation set results, the loss is 1.8094 for the first epoch and 0.6583 on the fiftieth epoch. The accuracy is 0.3431 on the first epoch and 0.7695 on the fiftieth epoch.

From the training set results of Adam model, the loss is 2.2084 for the first epoch and 1.1941 for the fiftieth epoch. The accuracy is 0.2406 for the first epoch and 0.5836 for the fiftieth epoch. From the validation set results, the loss is 1.7005 for the first epoch and 0.9746 on the fiftieth epoch. The accuracy is 0.3766 for the first epoch and 0.6624 for the fiftieth epoch.

From the result, POGD is significantly improved compared to Adam training. Contrast two algorithm training set results, the loss difference for the first epoch is 0.0822 but the difference for the fiftieth epoch is 0.4253. The accuracy difference for the first epoch is 0.0742 and for the fiftieth epoch is 0.1466. Due to the large learning rate used to maintain fairness, POGD has fluctuation in few epochs on the validation set. However, by the increase of epoch, this fluctuation is significantly reduced and becomes more stabilize.

In this experiment, Adam is convergence on the training set around 0.6, this also limits its performance on the validation set. Therefore, in the second experiment, Adam encountered a local minimum, which make it difficult for the model to continue learning. Under the same conditions, POGD can still show the trend of continuing learning and have better results.

\section{Conclusion}
Particle Optimized Gradient Descent is a new algorithm that integrates PSO and Adagrad gradient descent. It is achieved by proposing bold assumptions about the original principles and making adjustments based on the actual condition. Its position iteration is affected by the original position, acceleration and the position of the previous time. The velocity expectation starts from the retrospective value and predicted by global prediction and previous gradient. 

From the experience, POGD has fast value update speed, therefore it has good convergence speed. However, there need to adjust parameter carefully in each application to prevent over-learning or extreme weight changes. POGD decreases loss during training and also has adaptive learning ability. When the learning rate is set as a constant, the actual learning rate will gradually decrease with the iteration of POGD. The concept from PSO enables POGD to have good global optimization capability and better prevent from the local minimum.

To sum up, for the convolutional neural networks in image classification, when it is encountering slow convergence speed, global optimization of training and local minimum, POGD will be a good choice for the further experiments.


\clearpage
\appendix
\section{Appendix Figures}
\begin{figure}[ht]
	\centering
	\includegraphics[width=4.1in]{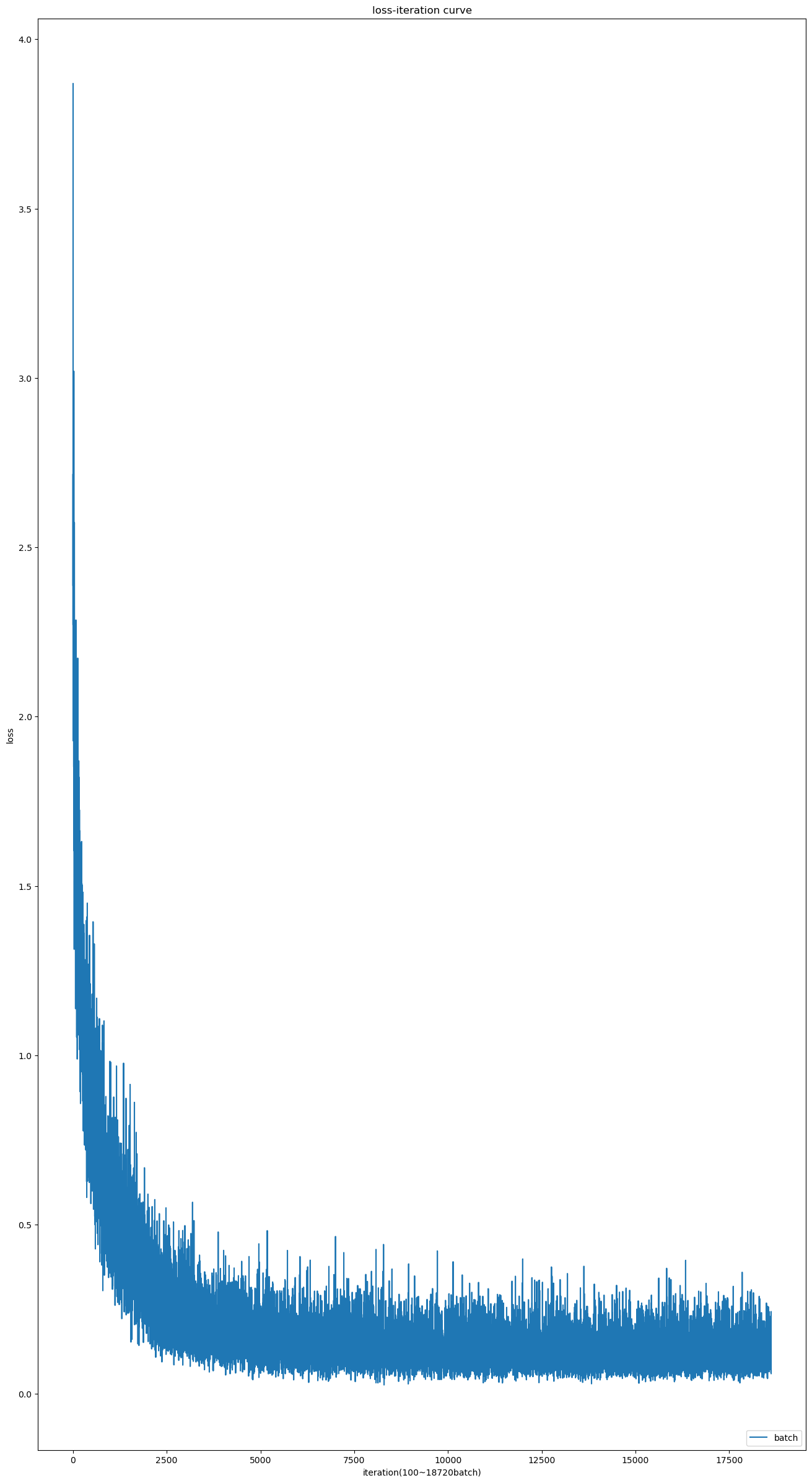}
	\caption{POGD Loss Iteration on Mnist Dataset}
	\label{fig_PLIM}
\end{figure}

\begin{figure}[ht]
	\centering
	\includegraphics[width=4.1in]{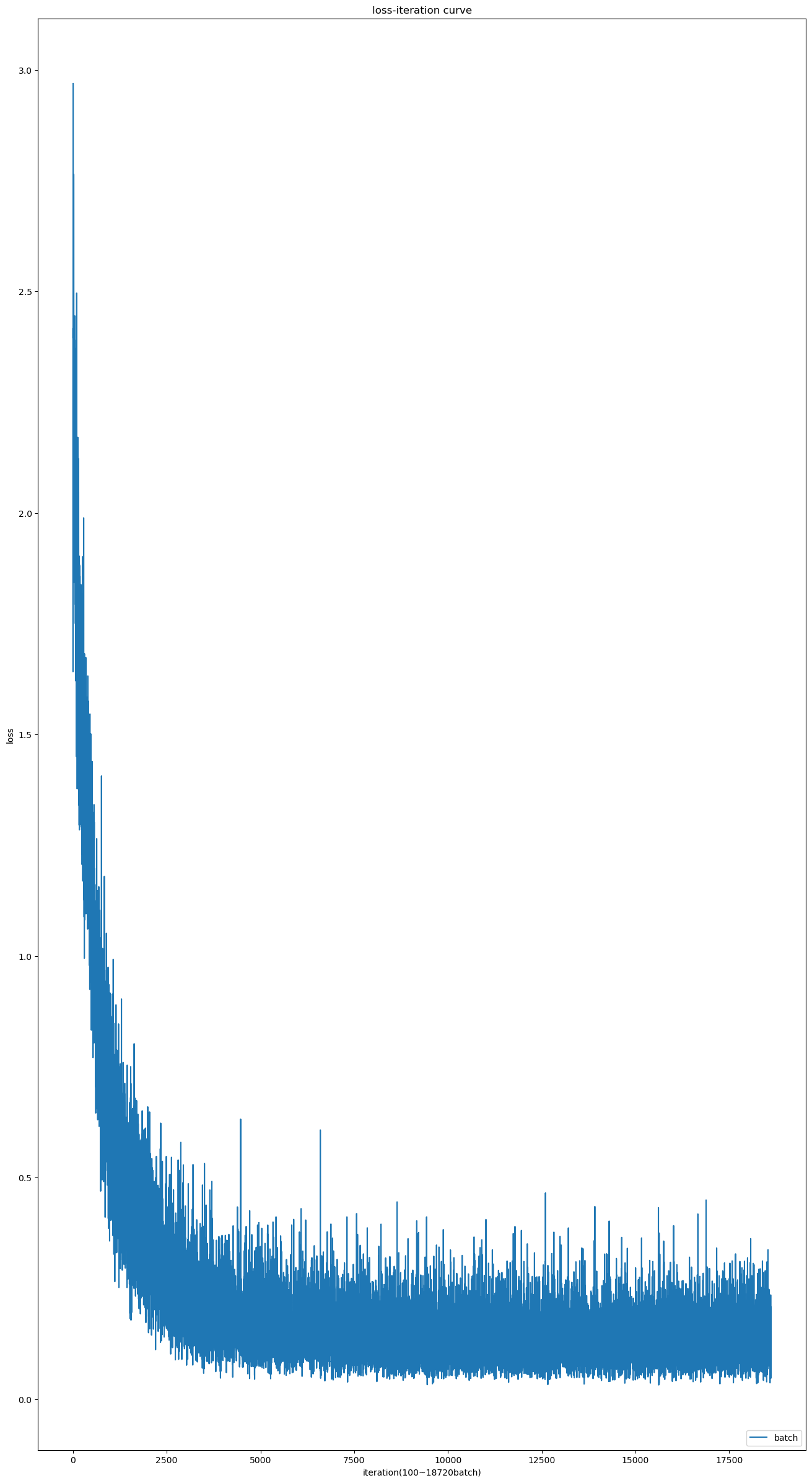}
	\caption{Adam Loss Iteration on Mnist Dataset}
	\label{fig_ALIM}
\end{figure}

\end{document}